\documentclass{article} 
\usepackage{iclr2023_workshop,times}

\usepackage{amsmath,amsfonts,bm}

\def\eqref#1{equation~\ref{#1}}

\def\1{\bm{1}}

\DeclareMathAlphabet{\mathsfit}{\encodingdefault}{\sfdefault}{m}{sl}
\SetMathAlphabet{\mathsfit}{bold}{\encodingdefault}{\sfdefault}{bx}{n}

\usepackage{hyperref}
\usepackage{url}
\usepackage{natbib}
\usepackage{forest}
\usepackage{multicol}
\usepackage{multirow}
\usepackage{comment}
\usepackage{algorithm}
\usepackage{algpseudocode}

\title{Symbolic Regression for PDEs using Pruned Differentiable Programs}

\iclrfinalcopy

\author{Ritam Majumdar, Vishal Jadhav \& Anirudh Deodhar\\
Tata Consultancy Services Research, India\\
\texttt{\{ritam.majumdar,vishal.jadhav,anirudh.deodhar\}@tcs.com} \\
\AND
Shirish Karande, Lovekesh Vig \& Venkataramana Runkana \\
Tata Consultancy Services Research, India \\
\texttt{\{shirish.karande,lovekesh.vig,venkat.runkana\}@tcs.com}
}

\begin{document}

\maketitle

\begin{abstract}
Physics-informed Neural Networks (PINNs) have been widely used to obtain accurate neural surrogates for a system of Partial Differential Equations (PDE). One of the major limitations of PINNs is that the neural solutions are challenging to interpret, and are often treated as black-box solvers. While Symbolic Regression (SR) has been studied extensively, very few works exist which generate analytical expressions to directly perform SR for a system of PDEs. In this work, we introduce an end-to-end framework for obtaining mathematical expressions for solutions of PDEs. We use a trained PINN to generate a dataset, upon which we perform SR. We use a Differentiable Program Architecture (DPA) defined using context-free grammar to describe the space of symbolic expressions. We improve the interpretability by pruning the DPA in a depth-first manner using the magnitude of weights as our heuristic. On average, we observe a 95.3\% reduction in parameters of DPA while maintaining accuracy at par with PINNs. Furthermore, on an average, pruning improves the accuracy of DPA by 7.81\% .  We demonstrate our framework outperforms the existing state-of-the-art SR solvers on systems of complex PDEs like Navier-Stokes: Kovasznay flow and Taylor-Green Vortex flow. Furthermore, we produce analytical expressions for a complex industrial use-case of an Air-Preheater, without suffering from performance loss viz-a-viz PINNs. 
\end{abstract}

\section{Introduction}

Symbolic Regression is the task of generating a mathematical expression that best fits a given dataset. SR is an important problem as it helps understand underlying relationships and patterns in data, with application in scientific discovery \cite{doi:10.1063/1.5116183,https://doi.org/10.1002/aic.17695}, engineering design \cite{article,doi:10.1063/1.5136351}, and financial forecasting \cite{Mostowfi2022}, just to name a few. SR helps reduce the complexity of the models and provides interpretable solutions, thereby improving the transparency and accountability of AI systems. Recently, Virgolin et al. in \cite{https://doi.org/10.48550/arxiv.2207.01018} proved SR to be an NP-hard problem. Historically, SR has been attempted using genetic programming methods \cite{https://doi.org/10.48550/arxiv.2205.09751}, purely Deep-learning methods like sequence generation \cite{DBLP:journals/corr/abs-2106-14131, DBLP:journals/corr/DevlinUBSMK17}, tree search \cite{DBLP:journals/corr/abs-1901-07714, https://doi.org/10.48550/arxiv.1905.11481, DBLP:journals/corr/abs-2006-10782}, and a combination of both Deep-learning and Genetic programming methods \cite{petersen2021deep, DBLP:journals/corr/abs-1801-03526}.

While SR has been applied for PDE equation discovery using Genetic Programming \cite{DBLP:journals/corr/abs-1903-08011}, Fast-function extraction \cite{article}, replacing activation functions of NNs with primitive functions \cite{https://doi.org/10.48550/arxiv.2207.00529}, sequence to sequence equation generation using Transformers \cite{DBLP:journals/corr/abs-1912-01412}, very few works \cite{DBLP:journals/corr/abs-2011-06673,https://doi.org/10.48550/arxiv.2207.06240} attempts to directly model the final analytical solution of the governing PDE. Inspired by \cite{NEURIPS2021_5c5a93a0} which generates differentiable programs, Majumdar et al. in \cite{https://doi.org/10.48550/arxiv.2207.06240} introduced Physics Informed Symbolic Networks (PISN) to generate analytical expressions for PDEs. Given context-free grammar, they approximate a production rule by taking a linear weighted approximation of the rules. While PISNs performed on par with PINNs, the analytical expressions generated were large and weren't interpretable. In this work, we use the original differentiable program architecture (DPA) in \cite{NEURIPS2021_5c5a93a0} for performing symbolic regression over generated data points by PINNs. We improve the transparency of the symbolic expressions by pruning the DPA in a depth-first manner, using the magnitude of weights as the heuristic. Pruning allows us to obtain sparser representations for PDEs that are easily interpretable.

Our key contributions are as follows: 1) We use a DPA to perform Symbolic Regression on PDEs. 2) Our pruning strategy reduces 95.3\% of the parameters of the program architecture with a performance  at par with PINNs. 3) Our framework demonstrates excellent performance on complex PDEs like Navier Stokes and industrial systems like Air-preheater \cite{PHM,https://doi.org/10.48550/arxiv.2212.10032} which have no predefined analytical solution.

The rest of the paper is organized as follows. Section \ref{methodology} consists of Methodology, followed by Observations and Discussions. Section 4 consists of the limitations. The Appendix is organized as follows. Section \ref{PDE_details} consists of PDE-details, followed by Experiment details. Section \ref{ts} consists of training schedule, followed by generated symbolic expressions and a comparison between pruned and unpruned DPA. Finally, we conclude by providing a visualization of the proposed pruning algorithm. 

\section{Methodology}
\label{methodology}

\begin{algorithm}
\caption{Symbolic Regression for Partial Differential Equation}
\label{end-to-end}
\begin{algorithmic}[1]
\State Train a PDE solver to solve for the PDE 
\State Generate input-output data points using the learned PDE solver
\State Perform Regression on generated data points using Differentiable Program Architecture
\State Prune the Differentiable Program Architecture
\end{algorithmic}
\end{algorithm}

\begin{figure}
    \centering
    \begin{forest}
for tree={
circle,
draw,
minimum size=3mm}
[root
[sin,edge label={node[midway,left=1pt]{\textbf{w}}}
[sin[x][y][1]][exp[x][y][1]]]
[exp[sin[x][y][1]][exp[x][y][1]]]]
\end{forest}
\caption{Program Derivation Graph of a Grammar with operators: $sin$, $exp$, $x$, $y$, $c$ of Depth 2}
    \label{fig:DSFPA_d2}
\end{figure}
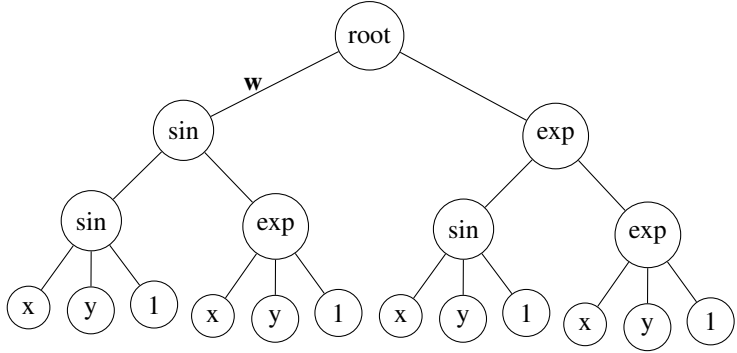

Algorithm \ref{end-to-end} describes our end-to-end procedure for performing symbolic regression on Partial Differential Equations. In the first step, we use a PDE solver to solve for the given PDE setup. In this work, we use PINNs as our PDE solver. The PINN solver can be replaced by any numerical PDE solver suitable for the problem at hand. The second step involves preparing the dataset by generating input-output data points using the trained PINN. Symbolic regression is then performed using a Differentiable Program Architecture defined based on the context-free grammar \cite{CFG} described in Equation \ref{CFG}. We take sin, exp, log, power 2, and power 3 as our unary operators, and Addition and Multiplication as our binary operators. Figure \ref{fig:DSFPA_d2} provides an example of expanding the differentiable program architecture till depth 2 using sin, exp, and leaf nodes as operators. $x$,$y$,$t$,$c$ are the terminal symbols. Finally, we prune the DPA as described in Algorithm \ref{DFS}.

\begin{equation}
\label{CFG}
\alpha::= \textbf{sin}\:\alpha_1\:|\:\textbf{exp}\:\alpha_1\:|\: 
\textbf{log}\:\alpha_1\:|\:\textbf{pow2}\:\alpha_1\:|\:\textbf{pow3}\:\alpha_1\:|\:\textbf{Add}\:\alpha_1\:\alpha_2\:|\: \textbf{Multiply}\:\alpha_1\:\alpha_2\:|\:x\:|\:y\:|\:t\:|\:c
\end{equation}

\begin{algorithm}
\caption{Depth First Search Pruning Strategy of Differentiable Program Architecture (DPA)}
\label{DFS}
\begin{algorithmic}[1]
\State $\textbf{Global initialize}$ $W \gets $ Unpruned DPA weights, loss $\gets$ \Call{Score\:}{$W$}
\State \textbf{Function} DFS ($node$)
\State $\textbf{Initialize}$ $visited \gets {node}$, $children \gets$ children of $node$ sorted by absolute value 
\If{$children$ is $None$}
\State $W^{'} \gets $ \Call{Prune\:}{$node$}
\State $W^{'} \gets $ \Call{Finetune\:}{$W^{'}$}
\State finetuned-loss $\gets$ \Call{Score\:}{$W^{'}$}
\If{finetuned-loss $\le$ loss}
\State loss $\gets$ finetuned-loss
\State $W \gets W^{'}$
\EndIf
\State {\textbf{Return} $W$,loss}
\EndIf
\ForAll{$child$ in $children$}
\If{$child$ not in $visited$}
\State \Call{DFS}{$child$}
\EndIf
\EndFor
\end{algorithmic}
\end{algorithm}

Algorithm \ref{DFS} represents our pruning strategy for DPA. We initialize the final DPA with unpruned DPA weights and loss as the mean-relative-L2-error on data points generated using the PDE solver. We postulate the importance of a term in the mathematical expression is directly proportional to the magnitude of the edge. Thus, we recursively visit every child of a node starting with the child having a minimum value, in a depth-first manner. On encountering a leaf, we prune that weight and finetune the DPA. If the resulting DPA performs on par or better, we accept the prune, else we reset the prune and move to the next child. We perform this operation recursively until all nodes are visited.

\section{Observations and Discussion}
\label{observations}

Table \ref{tab:Results} compares the performance of the output variables of interest. We compare the performance of pruned DPA with the trained PINN, unpruned DPA, and our benchmarks, AI-Feynmann (AIF), SymbolicGPT (SGPT), and Deep Symbolic Regression (DSR). Except for Air-Preheater (APH), we compare the relative-L2 error between the solutions generated with their corresponding true analytical solution. In the case of APH, true analytical solutions don't exist, and we use numerical simulations by Finite Difference Method to generate the ground-truth temperature distribution \cite{WANG201952}.

\begin{table}[h]
    \centering
    \begin{tabular}{|c|c|c|c|c|c|c|c|}
        \hline
        \vspace{1pt}
        &&PINN&DPA-Unpruned&DPA-Pruned&AIF&SGPT&DSR\\
        \hline
       Diffusion  & u & 7.32e-3&8.54e-3&\textbf{8.16e-3}&\textbf{8.00e-4}&0.54&1.16\\
       \hline
        \multirow{3}{*}{Kovasznay} &  u & \textbf{9.64e-3}&1.15e-2&\textbf{9.85e-3}&4.50e-1&0.53&0.64\\
         & v& \textbf{1.44e-2}&1.56e-2&\textbf{1.47e-2}&4.80e-1&0.55&0.54\\
        & p & 1.75e-2&2.09e-2&\textbf{1.77e-2}&\textbf{1.69e-2}&0.75&1.33\\
        \hline
        {Taylor}& u & \textbf{3.02e-2}&3.71e-2&\textbf{3.08e-2}&5.60e-1&0.59&1.52\\
         Green& v & \textbf{2.73e-2}&3.59e-2&\textbf{2.76e-2}&6.70e-1&0.62&1.83\\
        & p & \textbf{3.64e-2}&4.66e-2&\textbf{3.75e-2}&7.40e-1&0.76&0.91\\
        \hline
        Diffusion & \multirow{2}{*}{u} & \multirow{2}{*}{\textbf{1.34e-2}}&\multirow{2}{*}{1.68e-2}&\multirow{2}{*}{\textbf{1.64e-2}}&\multirow{2}{*}{3.50e-1}&\multirow{2}{*}{0.56}&\multirow{2}{*}{0.82}\\ 
        Reaction &&&&&&&\\ 
        \hline
        \multirow{6}{*}{APH}&\(T_{fg}\)&\textbf{2.03}&2.14&\textbf{2.06}& 31.93&35.21&22.45\\
        &\(T_{mg}\) &\textbf{2.53}&2.55&\textbf{2.54}& 34.23&38.24&31.27\\
        &\(T_{fa_1}\)&\textbf{3.08}&3.27&\textbf{3.09}& 28.45&46.65&44.48\\
        &\(T_{ma_1}\)&\textbf{2.81}&3.02&\textbf{2.85}& 47.71&41.24&49.67\\
        &\(T_{fa_2}\) &\textbf{2.97}&3.10&\textbf{3.00}&39.64&37.41&45.21\\
        &\(T_{ma_2}\) &\textbf{3.02}&3.08&\textbf{3.04}&38.26&41.69&37.73\\
        \hline 
    \end{tabular}
    \caption{Performance comparison of Differentiable program architecture with existing benchmarks. In every row, bold denotes the Top-2 best performing methods.}
    \label{tab:Results}
\end{table}

Across all tasks, Unpruned-DPA is slightly worse than PINNs on the same training samples, in spite of the best expression lying in the defined CFG. The reason is, NNs have one standard activation throughout, while DPA has multiple operators with unique convergence characteristics. For example, Sine introduces periodicity, leading to the gradients being periodic, exponential increases rapidly for larger values and log decreases sharply for very small values. Currently, popular optimizers like Adam have difficulty in converging all the operators simultaneously \cite{8407425}, as they all have different convergence rates. Loss-curve characteristics of DPA need to be studied and custom optimizers have to be developed further. Across all examples, we observe pruning to improve the accuracy of DPA. The accuracy boost happens because pruning of weights reduces overfitting occurring due to complex expressions at higher depths, leading to better generalization. From table \ref{tab:Pruned_characteristics}, we observe on an average, 95.3\% reduction in \(\#\) weights of DPA after pruning. Furthermore, pruned DPA comprehensively outperforms the benchmarks. Across the 14 output variables in 5 PDEs, AIF is marginally better on just 2 variables whose underlying mathematical expressions are Depth-1 expressions and are easy to capture. The other 12 variables have expressions with higher depths which the benchmarks struggle to recapture. Failure of benchmarks on expressions with higher depth highlights the superior representation capacity of program architecture. 

Table \ref{tab:Pruned_Expressions} and \ref{tab:Benchmark_expressions} represent the expressions obtained by Pruned-DPA and benchmark methods respectively. On careful observation, we notice SGPT and DSR to be biased towards the logarithm and nested sin and cos operators. AIF produces over-simplified expressions, hence struggles to find a good fit to datapoints. We provide an example in Appendix \ref{pruned} of unpruned-DPA for the Diffusion equation. Our pruning heuristics allows us to prune 95\% of the weights and bring down 20 line expressions to 1-2 lines,  drastically improving their explainability. Furthermore, for complex examples like Kovasznay flow, pruned-DPA expressions come very close to the ground-truth, and equivalencies can be proved. \(u_{true}\) and \(u\) differ in \(cos(2\pi y)\) and \(sin(6.28y-1.57)\) terms, and it's well known from trignometric identities relation, \(cos(2\pi y)=sin(2\pi y-pi/2)\). \(v_{true}\) and \(v\) differ in \(exp(\lambda x)/2\pi\) and \((0.29-0.54x+0.46x^2-0.28x^3+0.12x^4)\), where one can verify, the differed expression is in fact the Taylor-series approximation of the ground-truth upto the \(4^{th}\) order. 

The higher relative L2-error of PINNs and DPA in Taylor Green Vortex is because of the difficulty in the underlying physics dynamics, as the flow is unsteady with decaying vortices \cite{10.1093/imanum/drac085}. Here, the advantages of pruning are more evident, as there is a 0.79\% boost in accuracy over pruned-DPA, indicating significant improvement in function generalization arising from disentanglement from complex expressions, thereby improving interpretability. In the Diffusion-Reaction example, the ground-truth is an example of Depth-6. We were able to obtain a symbolic expression using a Depth-3 DPA, highlighting the expressive capacity of DPA. However, it comes at a cost of larger complex expressions with terms like log, \(2^{nd}\) and \(3^{rd}\) power terms which don't appear in the ground truth. In contrast to other examples, there isn't a large boost in accuracy from pruning here, as there are large inter-dependencies amongst operators to fit a Depth-6 expression into a Depth-3 DPA. Our framework shows promise for systems with no-analytical solution, as evident from the example of Air-Preheaters, where benchmark methods fail. Pruned-DPA gives 1-line symbolic expressions for generalizing temperature distributions in the entire domain, while the temperature MAE w.r.t. numerical simulations are comparable to that of PINNs.

\section{Limitations}
\label{Limitations}
While our pruning strategy reduces the size of DPA, the ground-truth expressions are far more concise than the obtained expressions. One of the reasons is the greedy nature of the pruning algorithm. To provide an intuition, suppose a lesser weight is assigned to the operator when the DPA hasn't been pruned yet, which can ultimately provide the most concise expression. There is a high possibility of that weight getting pruned, as a different expression obtained by the remaining operators can still generalize over the dataset. Thus, our pruning strategy is sub-optimal in nature, and better pruning strategies need to be explored to obtain even more concise expressions. Nevertheless, this DFS-based pruning is a good starting point. Additionally, the convergence of DPA is non-trivial due to the varying mathematical properties of primitive operators. Detailed investigation on optimization guarantees of DPA and theoretical studies on its convergence and error bounds remain.

\newpage

\bibliography{iclr2023_workshop}

\appendix

\section{Appendix}

\subsection{PDE information}
\label{PDE_details}

Table \ref{tab:PDE_information} represents the PDE setups for our experiments. The second column consists of information on governing conditions, initial conditions, boundary conditions, and the domain of the spatial and temporal variables. The final column represents the ground-truth analytical solutions of output variables of interest.  

\begin{table}[h]
    \centering
    \begin{tabular}{|c|c|c|}
        \hline
        &Governing Conditions&Ground-truth expressions\\
        \hline
       \multirow{3}{*}{Diffusion}&\(u_t = u_{xx} - e^{-t}sin(\pi x)(1-\pi^2)\)&\\
       &\(u(x,0)=sin(\pi x)\)&\(u_{true}=e^{-t}sin(\pi x)\) \\
       & \(u(-1,t)=u(1,t)=0\)&\\
       \hline
        \multirow{3}{*}{Kovasznay}&\(u\cdot\nabla u+\nabla p = \nu\Delta u\) \;\;\;\; in $[0,1]^2$&\(u_{true}= 1-e^{\lambda x}cos(2\pi y)\)\\
         &\(div(u)=0\) \;\;\;\;\;\;\;\;\;\;\;\;\;\;\;\;\;\; in [0,1]&\(v_{true}=\lambda e^{\lambda x}sin(2\pi y)/2 \pi\) \\
        &&\(p_{true}= (1-e^{2\lambda x})/2\)\\
        \hline
        \vspace{2pt}
        \multirow{3}{*}{Taylor-Green}&\(u_t+u\cdot\nabla u+\nabla p = \nu\Delta u\) \;\; in $[0,2]^2$$\times$[0,1]&\(u_{true}=-cos(\pi x)sin(\pi y)e^{-2\pi^2\nu t}\)\\
         &\(div(u)=0\) \;\;\;\;\;\;\;\;\;\;\;\;\;\;\;\;\;\;\;\;\;\;\;\; in $[0,2]^2$$\times$[0,1]&\(v_{true}=sin(\pi x)cos(\pi y)e^{-2\pi^2\nu t}\)\\
        & \(u(t = 0) = u_0\) \;\;\;\;\;\;\;\;\;\;\;\;\;\;\;\;\;\;\; in $[0,2]^2$ &\(p_{true}=-\frac{(cos(2\pi x)+cos(2\pi y))e^{-2\pi^2\nu t}}{4}\)\\
        \hline
        \vspace{2pt}
        &\(u_t = u_{xx} + e^{-t}f(x)\)&\(u_{true}=e^{-t}(p(x)+q(x))\)\\ 
        Diffusion&\(u(x,0)=p(x)+q(x)\)&\(p(x)=\frac{12sin(x)(1+cos(x))+4sin(3x)}{12}\)\\
        Reaction&\(u(t,-\pi)=u(t,\pi)=0\) \;\;\;\;\;\; in [$-\pi$,$\pi$]$\times$[0,1]&\(q(x)=\frac{sin(4x)(1+cos(4x)}{4}\)\\
        &\(f(x)=\frac{36sin(2x)+64sin(3x)+90sin(4x)+189sin(8x)}{24}\)&\\
        \hline 
    \end{tabular}
    \caption{PDE-information}
    \label{tab:PDE_information}
\end{table}

\begin{figure}[h]
    \centering
    \includegraphics[height=5cm, width=\textwidth]{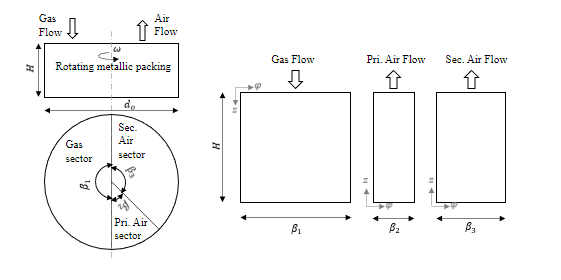}
    \caption{Air-Preheater schematic derived from \cite{https://doi.org/10.48550/arxiv.2212.10032}. The computational domain is divided into three parts, Gas, primary air and secondary air respectively.}
    \label{fig:APH}
\end{figure}

\textbf{Air-Preheater:} We consider the non-dimensional form of APH. Equation \ref{eq:cond} represents Conduction while \ref{eq:conv} represents convection heat transfer. There are six outputs to this PDE system, three fluid temperatures ($T$) and three metal temperature ($T_m$) for given co-ordinates ($\theta,z$). $NTU$ and $Pe$ stand for the number of transfer units and Peclet number respectively.

\begin{equation}
\label{eq:cond}
    \frac{\partial T_{m_{j}}}{\partial\varphi} = NTU_{m_{j}} (T_j - T_{m_j}) + \frac{1}{Pe_{m_j}}   \frac{\partial^2 T_{m_j}}{\partial z^2} 
\end{equation}
\begin{equation}
\label{eq:conv}
    \frac{\partial T_j}{\partial z} = {NTU}_{m_j} \left( {T}_{m_{j}} - {T}_{j}, \right) \hspace{0.2cm} j = 1,2,3
\end{equation}
\begin{equation}
    \label{eq:bndcnd_gas}
    {T_j}\left( \varphi, z=0 \right) = {T}_{in,j},  \hspace{0.2cm} j = 1,2,3
\end{equation}
\begin{equation}
    \label{eq:intface_gas}
    T_{m_1}(\varphi = 0, z) = T_{m_3} (\varphi = 1, 1-z)
\end{equation}
\begin{equation}
    \label{eq:intface_priair}
    T_{m_1}(\varphi = 1, z) = T_{m_2} (\varphi = 0, 1 - z)
\end{equation}
\begin{equation}
    \label{eq:intface_secair}
    T_{m_2} (\varphi = 1, z) = T_{m_3} (\varphi = 0, z) 
\end{equation}
\begin{equation}
    \label{eq:gradcnd}
    \frac{\partial {T}_{m_j}[z = 0, 1]}{\partial z} = 0,  \hspace{0.2cm}j = 1,2,3 
\end{equation}

The boundary conditions are imposed by Gas inlet temperature ($T_{in,1}$), primary air inlet temperature ($T_{in,2}$), and secondary air inlet temperature ($T_{in,3}$) in Equation \ref{eq:bndcnd_gas}. Equations \ref{eq:intface_gas},\ref{eq:intface_priair},\ref{eq:intface_secair} impose
continuity constraints on the metal temperature.

\subsection{Experiments}
\label{Experiments}

We consider five systems of PDEs for our experiments, Diffusion equation \cite{https://doi.org/10.1002/andp.18551700105}, Navier-Stokes: Kovasznay flow \cite{https://doi.org/10.1002/andp.18551700105}, Navier-Stokes: Taylor Green Vortex equation \cite{10.1093/imanum/drac085}, Diffusion Reaction equation, and two dimensional conjugate heat transfer in Air-Preheater\cite{WANG201952}.  Diffusion-Reaction PDEs are important for modeling chemical reactions \cite{SHEU2000123} wherein there is a formation of new chemical products, and diffusion wherein there is a transfer of matter over a domain. Kovasznay flow is a two-dimensional steady-state Navier-Stokes equation with Reynold's Number of 20. Taylor-Green Vortex flow is a two-dimensional unsteady Navier-Stokes equation with viscosity $\nu=0.01$. For both Kovasznay flow and Taylor-Green Vortex, we sample the boundary conditions from the ground-truth analytical solutions. Our final use case is that of Air-Preheaters (APH). APH is a heat exchanger deployed in thermal power plants to improve the thermal efficiency. Monitoring of internal temperature profiles of APH is important to avoid failures, which arises due to complex thermal and chemical phenomena. The reference solution of APH is derived using a Finite-Difference method and doesn't have a ground-truth analytical solution. Inspection of internal temperature profiles can significantly benefit from symbolic representations in contrast to NNs due to improved interpretability. We describe the schematics of APH in \ref{PDE_details} and governing equations of other PDE systems in Table \ref{tab:PDE_information}. 

\subsection{Training Schedule}
\label{ts}

\textbf{PINNs:} We consider a Neural Network with 3 hidden layers of 50 neurons each. We train using Adam optimizer for 60k epochs with multiplicative decay of 0.1 every 15k epochs, starting from a learning rate of $1e^{-3}$. We further finetune using the L-BFGS optimizer. In the examples of Diffusion, Diffusion Reaction, there are two input variables, $x,t$. We consider a computational domain of $[0,1]$ \textbf{X} $[0,1]$ with 10201 collocation points divided into 7500 training points and 2701 test points. In Kovasznay flow, we consider a 101x101 equally spaced grid domain to represent the XY plane, where X, Y $\in$ [-0.5,1.0] \textbf{X} [-0.5,1.5]. We consider 2601 collocation points and 320 boundary condition points, with 80 points for each face of the grid. Taylor-Green Vortex: the computational domain is defined for X,Y,T as [0,2] \textbf{X} [0,2] \textbf{X} [0,1], with a time-step of 0.1 in the time dimension. We sample 25k collocation points, 5k initial-boundary condition points throughout the domain. For evaluating the test-performance, across all PDE examples, we sample 10k points for every output variable of interest. 

\textbf{Differentiable Program Architecture}: Table \ref{tab:Pruned_characteristics} consists of the information on the maximum depth of the architecture and operators of interest for every PDE system. The weights of the DPA are initialized using the Glorot-uniform optimizer which is used to perform SR. We use an Adam Optimizer for 100k epochs with multiplicative decay of 0.1 every 25k epochs with an initial learning rate of $1e^{-2}$. We further apply early stopping if the generalization on datapoints generated by PINNs doesn't improve in the last 5k epochs. For training the unpruned-DPA, we apply L1-regularization over all weights with L1-coefficient of $1e^{-5}$. All experiments were conducted on Nvidia P100 GPU with 16 GB GPU Memory and 1.32 GHz GPU Memory clock using Pytorch framework. 

\textbf{SymbolicGPT:} The training methedology of SymbolicGPT is adapted from the original paper \cite{DBLP:journals/corr/abs-2106-14131}. The hyperparameter specifications are as follows: numEpochs: 20, embeddingSize: 512, number of points:10k, blockSize: 200, testBlockSize: 400, batchSize: 128, variable-embedding: False.

\textbf{AI-Feynman:} The training methodology of AI-Feynman is adapted from the original paper \cite{DBLP:journals/corr/abs-2006-10782}. We consider the following hyperparameter sets for our experiments: 'bftt':\{60,120\}, 'epochs':\{300,400,500\}, 'op':\{'7ops.txt','10ops.txt','14ops.txt','19ops.txt'\}, 'polynomial degree':\{3,4,5\}.  

\textbf{DSR:} The training methodology of Deep Symbolic Regression is adapted from the original paper \cite{petersen2021deep}. We consider the following hyperparameter sets for our experiments: 'seed':\{1,2,3,4,5\}, 'function set':\{'add', 'sub', 'mul', 'div', 'sin', 'cos', 'exp', 'log', 'const'\}, batch size:1000, learning rate: $5e^{-4}$, entropy coefficient: 0.005, risk factor: 0.05. 

\begin{table}[h]
    \centering
    \begin{tabular}{|c|c|c|c|c|c|}
        \hline
        \vspace{1pt}
        &&Depth&Unpruned&Pruned&Operators\\
        \hline
       Diffusion&u&2&157&15&[$x$,$t$,$sin$,$exp$,$+$,$*$]\\
       \hline
        \multirow{3}{*}{Kovasznay} & u & 2& 343&12&[$x$,$y$,$sin$,$exp$,$+$,$*$,$log$,$pow2$,$pow3$]\\
         & v& 2& 343&13&[$x$,$y$,$sin$,$exp$,$+$,$*$,$log$,$pow2$,$pow3$]\\
        & p & 2& 343&9&[$x$,$y$,$sin$,$exp$,$+$,$*$,$log$,$pow2$,$pow3$]\\
        \hline
        \multirow{3}{*}{Taylor-Green}&u&3&3097&20&[$x$,$y$,$t$,$sin$,$exp$,$+$,$*$,$log$,$pow2$,$pow3$]\\
         &v& 3&3097&33&[$x$,$y$,$t$,$sin$,$exp$,$+$,$*$,$log$,$pow2$,$pow3$]\\
        &p& 3&3097&35&[$x$,$y$,$t$,$sin$,$exp$,$+$,$*$,$log$,$pow2$,$pow3$]\\
        \hline
        Diffusion & \multirow{2}{*}{u}&\multirow{2}{*}{3}&\multirow{2}{*}{3097}&\multirow{2}{*}{63}&\multirow{2}{*}{[$x$,$t$,$sin$,$exp$,$+$,$*$,$log$,$pow2$,$pow3$]}\\ 
        Reaction &&&&&\\ 
        \hline
        \multirow{6}{*}{Air-Preheater}&\(T_{fg}\)& 2&157&6&[$\theta$,$z$,$sin$,$exp$,$+$,$*$]\\
        &\(T_{mg}\)& 2&157&15&[$\theta$,$z$,$sin$,$exp$,$+$,$*$]\\
        &\(T_{fa_1}\)& 2&157&15&[$\theta$,$z$,$sin$,$exp$,$+$,$*$]\\
        &\(T_{ma_1}\)& 2&157&9&[$\theta$,$z$,$sin$,$exp$,$+$,$*$]\\
        &\(T_{fa_2}\) & 2&157&8&[$\theta$,$z$,$sin$,$exp$,$+$,$*$]\\
        &\(T_{ma_2}\) & 2&157&12&[$\theta$,$z$,$sin$,$exp$,$+$,$*$]\\
        \hline 
    \end{tabular}
    \caption{Differentiable Program Architecture characteristics. Unpruned and pruned refers to $\#$ parameters in the resulting DPA respectively.}
    \label{tab:Pruned_characteristics}
\end{table}

\newpage

\subsection{Generated Symbolic expressions}

\begin{table}[h]
    \centering
    \begin{tabular}{|c|l|}
        \hline
        \vspace{1pt}
       \multirow{2}{*}{Diffusion}&\(u=(1.51x-2.04sin(-2.51x+0.20t))(2.62x+0.32t)\)\\
       &\(+3.67sin(1.30sin(0.37t+1.63)+3.13x)-3.53\)\\
       \hline
        &\(u=1.01+0.99(sin(6.28y-1.57)e^{-1.81x})\)\\
        Kovasznay& \(v=sin(6.28y-3.14)(0.29-0.54x+0.46x^2-0.28x^3+0.12x^4)\) \\
        Flow& \(p=-2x^4+4.38x^3-3.75x^2+1.81x +0.02\)\\
        \hline
        & \(u=(1.93x^2-0.13xt-0.03t-0.26x-0.36)(4.13y-4.12)*\)\\
        &\((-0.72sin(1.71y)+xy-0.56y^2+0.06y)+0.50\)\\
        Taylor-Green& \(v=0.21sin(-0.33xt+0.34yt+3.41x-3.43y)+(0.74x+0.54y+0.13t-1.09)\)\\
        Vortex&\(*(-1.03x-0.93y-0.31)(0.31x-0.69sin(1.69x+1.87y-2.11))\)\\
        & \(p=0.75sin(h_1)-0.29h_3-0.17(sin((-0.91x+0.32t+0.61)+3.30))\)\\
    &\(h_3 = sin(2.07(-0.18t+1.84)(-1.64x-0.44t+0.41))\)\\
    &\(h_1 = 0.25t-1.61sin(1.89y+0.08)+(1.30y-0.03t)(0.18y+0.23t-1.41)+0.33\)\\
        \hline
         & \(u = 0.30log(0.64log(-0.44x^2-1.9x-0.95))-0.07h_2^2+h_3\)\\
        &\(h_2 = 0.84-0.25sin(0.95x-0.33)+0.46h_8^2-0.43h_9^3\)\\
        Diffusion&\(h_3 = 0.19(-0.26x-0.12y-0.15e^{-0.66-0.22y})^3\)\\
        Reaction&\(h_8 = 0.87x-0.47e^{0.83x+0.64}+0.83(-0.99x+0.33y-0.25)\)\\
        &\((0.49x+0.02y+0.71)-0.07(0.07x-0.30y-0.04)^3\)\\
        &\(h_9 = 0.17sin(0.22x-0.07y-0.28)-0.19(0.67x-0.14y-0.07)^2\)\\
        &\(-0.09(0.50x+0.72)^3+0.22x-0.07y-0.02\)\\
        \hline   
        \vspace{2pt}
         \multirow{6}{*}{APH}& \(T_{fg}=0.99sin(0.22z+0.83e^{-0.21z+0.61\theta+0.66})\)\\
         & \(T_{mg}=(0.52sin(1.56\theta)+1.18)(0.83e^{-0.44\theta}+0.18z\theta-0.23\theta^2-0.27\theta)\) \\
        & \(T_{fa_1}=(-0.11z+0.77\theta-1.59+0.27(z\theta))(0.03z+0.22\theta-0.60+0.37\theta^2)\)\\
        & \(T_{ma_1}=(-1.14sin(0.07z+0.88\theta-1.06))(0.06z+0.69\theta+0.97)\)\\
        &\(T_{fa_2}=0.86sin(0.34z+1.26\theta+0.55)(-0.10z-1.02\theta)(0.11z+1.10\theta)+0.96\)\\
        & \(T_{ma_2}=(0.13z+1.30sin(1.08\theta)+1.25)(0.08z-0.68\theta+0.77)\)\\
        \hline
    \end{tabular}
    \caption{Expressions obtained after pruning DPA}
    \label{tab:Pruned_Expressions}
\end{table}

\begin{table}[h]
    \centering
    \begin{tabular}{|c|c|c|c|}
        \hline
        \vspace{1pt}
        &AI-Feynman&Symbolic GPT&DSR\\
        \hline
        u&\(1.01exp(-0.99t)*cos(0.99x-1.56)\)&\(1.22x-log(x-\theta)\)&\(sin(cos(sin(log(x)))\)\\
        \hline 
        u&\(1.24cos(x)\)&\(exp(0.43x-0.71y)+2.23\)&\(sin(sin(x))+tan(y)\)\\
        v&\(sin(x)-1.14cos(y)\)&\(1.52\)&\(2.57tan(y)\)\\
        p&\(0.51-0.48e^{\lambda x}\)&\(log(y-exp(1.13x)+0.47)\)&\(0.52x-1.62y\)\\
        \hline
        u&\(0.75tan(1.61sin(y-z)-0.33cos(t)\)&\(1.1704\)&\(1.6321\)\\
        v&\(4.43\)&\(1.82+1.10y-0.43t\)&\(x^3-1.22cos(t)-1.15\)\\
        p&\(2.24x-(1.19log(x)-0.47t-0.22)^3\)&\(0.95x-0.18y+1.44\)&\(0.08log(1.87)\)\\
        \hline
        u&\(1.27exp(x-t)(cos(x)-sin(cos(t)))\)&\(3.33e^{-t}(1.14x-x^2)\)&\(sin(sin(x))cos(cos(t))\)\\
        \hline
        \(T_{fg}\)&\(1.45log(2z\theta+(sin(z-\theta)^3)\)&\(\theta-log(0.96z+0.34)\)&\(sin(sin(sin(\theta)))\)\\
\(T_{mg}\)&\(2.67exp(1.33cos(\theta-5.53))-3.41\)&\(z-\theta^2\)&\(3.32\)\\
\(T_{fa_1}\)&\(tan(0.77z)-\theta*cos(z)-1.05\)&\(3.22log(1.19z)-0.46\)&\(cos(cos(cos(\theta)))\)\\
\(T_{ma_1}\)&\(1.22z^2\)&\(1.17\theta-0.95\)&\(z-\theta+cos(cos(1.32))\)\\
\(T_{fa_2}\)&\(1.64log(2.32\theta)-0.19z^3\)&\(2.43log(\theta)log(z)+3.21\)&\(cos(cos(1.33sin(z)-\theta))\)\\
\(T_{ma_1}\)&\(z-\theta\)&\(1.52log(\theta z)-0.96\)&\(1.67\theta-z\)\\
   \hline

    \end{tabular}
    \caption{Symbolic Expressions generated by benchmark methods}
    \label{tab:Benchmark_expressions}
\end{table}

\newpage
\subsection{Pruned Vs Unpruned: Diffusion Equation}
\label{pruned}

$U_{unpruned}=-1.76e^{-6}x+0.30t-0.78sin(h_{20})-0.43e^{h_{21}}-4.84e^{-5}(h_{22}+h_{23})+1.04h_{24}h_{25}+3.45e^{-5}\\
h_{20} = -0.57x-6.45e^{-5}t-0.54sin(h_{10})-0.47e^{h_{11}}+1.28(h_{12}+h_{13})+1.05e^{-4}h_{14}h_{15}+0.15\\
h_{21} = -1.16e^{-5}x-2.18e^{-5}t+7.28e^{-5}sin(h_{16})-1.85e^{-5}e^{h_{17}}-1.77e^{-4}(h_{18}+h_{19})-2.12e^{-5}h_{110}h_{111}+0.07\\
h_{22} = -1.16e^{-5}x-2.18e^{-5}t+7.28e^{-5}sin(h_{112})-1.85e^{-5}e^{h_{113}}-1.77e^{-4}(h_{114}+h_{115})-2.12e^{-5}h_{116}h_{117}-6.65e^{-5}\\
h_{23} = 7.42e^{-5}x+1.63e^{-5}t-1.60e^{-4}sin(h_{118})-8.29e^{-5}e{h^{119}}-6.21e^{-5}(h_{120}+h_{121})+1.59e^{-4}h_{122}h_{123}+0.0002\\
h_{24} = 6.04e^{-1}x-1.95e^{-1}t-8.08e^{-1}sin(h_{124})+7.38e^{-5}e^{h_{125}}-2.00e^{-1}(h_{126}+h_{127})-6.44e^{-5}h_{128}h_{129}+0.20\\
h_{25} = 3.39e^{-2}x-1.44e^{-5}t-8.03e^{-5}sin(h_{130})+3.85e^{-1}e^{h^{131}}-3.38e^{-5}(h_{132}+h_{133})+1.14h_{134}h_{135}+0.29\\
h_{10} = -0.04x+1.32t+0.60,\quad\quad\quad\quad\quad\quad\quad\quad\quad\quad h_{11} = 4.65e^{-1}x+6.25e^{-6}t+0.5031\\
h_{12} = -7.07e^{-1}x+1.47e^{-5}t+8.60e^{-6},\quad\quad\quad\quad\quad\quad h_{13} = -1.02x+1.49e^{-5}t+1.23e^{-5}\\
h_{14} = -0.0001x-0.0001t-0.0002,\quad\quad\quad\quad\quad\quad\quad h_{15} = -1.25e^{-4}x+5.77e^{-5}t-0.0002\\
h_{16} = -4.31e^{-5}-1.34e^{-4}-8.35e^{-5},\quad\quad\quad\quad\quad\quad\quad h_{17} = -1.64e^{-4}x+2.97e^{-5}t-7.37e^{-5}\\
h_{18} = 4.88e^{-6}x+7.52e^{-5}t-8.31e^{-5},\quad\quad\quad\quad\quad\quad\quad h_{19} = -9.6106e^{-5}x-9.97e^{-5}t+1.88e^{-5} \\
h_{110} = 5.38e^{-5}x-6.86e^{-5}t+0.0001,\quad\quad\quad\quad\quad\quad\quad h_{111} = -2.50e^{-4}x+9.12e^{-5}t-0.0001\\
h_{112} = 0.0001x+0.0001t-5.28e^{-5},\:\:\quad\quad\quad\quad\quad\quad\quad h_{113} = 1.23e^{-5}x+6.02e^{-5}t+2.20e^{-5}\\
h_{114} = 0.0002x-0.0002t+0.0003,\quad\quad\quad\quad\quad\quad\quad\quad h_{115} = 3.81e^{-5}x+7.75e^{-5}t+8.39e^{-5}\\
h_{116} = 3.08e^{-4}x+1.16e^{-5}t-0.0003,\quad\quad\quad\quad\quad\quad\quad h_{117} = 6.94e^{-5}x-1.38e^{-4}t-4.28e^{-5}\\
h_{118} = 9.81e^{-5}x-6.65e^{-5}t+3.90e^{-5},\quad\quad\quad\quad\quad\quad\quad h_{119} = 9.98e^{-5}x-1.34e^{-4}t+8.06e^{-5}\\
h_{120} = -4.02e^{-6}x+5.20e^{-5}t-0.0003,\quad\quad\quad\quad\quad\quad\quad h_{121} = -4.28e^{-5}x-1.22e^{-4}t-7.00e^{-6}\\
h_{122} = -0.0002x+0.0001t+9.46e^{-5},\quad\quad\quad\quad\quad\quad\quad\quad h_{123} = 4.80e^{-5}x-4.22e^{-5}t+3.66e^{-5}\\
h_{124} = -2.06x+0.65t+0.53,\quad\quad\quad\quad\quad\quad\quad\quad\quad\quad\quad h_{125} = 5.84e^{-5}x+1.66e^{-4}t-1.68e^{-5}\\
h_{126} = -2.03e^{-1}x+1.86e^{-5}t-2.52e^{-5},\quad\:\:\quad\quad\quad\quad\quad\quad h_{127} = -4.98e^{-6}x+1.8e^{-5}t+1.07e^{-5}\\
h_{128} = -3.32e^{-5}x-8.32e^{-5}t-0.0001,\:\:\quad\quad\quad\quad\quad\quad h_{129} = 2.89e^{-5}x+6.21e^{-5}t+0.0001\\
h_{130} = -1.42e^{-4}x+1.01e^{-7}t+6.28e^{-5},\quad\quad\quad\quad\quad\quad h_{131} = 3.57e^{-5}x-5.78e^{-2}t+0.37\\
h_{132} = -2.34e^{-4}x+1.66e^{-6}t-6.33e^{-5},\quad\quad\quad\quad\quad\quad h_{133} = 0.0001x+0.0002-0.0002\\
h_{134} = 0.81x-0.03t-0.65,\quad\quad\quad\quad\quad\quad\quad\quad\quad\quad\quad\quad h_{135} = -1.23x+0.11t+1.03e^{-5}\\
$

\(U_{pruned}=(1.51x-2.04sin(-2.51x+0.20t))(2.62x+0.32t)+3.67sin(1.30sin(0.37t+1.63)+3.13x)-3.53\)\\

\newpage
\subsection{Pruning Algorithm visualization}

As an example, we take a DPA of depth 2 with $sin$,$exp$,$log$ as operators, and $x$,$y$,$c$ as leaf nodes.
\begin{figure}[h]
    \centering
    \begin{forest}
for tree={
circle,
draw,
minimum size=2mm,
}
[root
[sin,edge label={node[midway,left=2pt]{-0.34}}
[sin][exp][log]]
[exp,edge label={node[midway,left]{\textbf{0.16}}}, edge={ultra thick}
[sin,edge label={node[midway,left]{-1.37}}]
[exp,edge label={node[midway]{1.24}}]
[log,edge label={node[midway,right]{\textbf{-0.05}}}, edge={ultra thick}
[x,edge label={node[midway,left]{1.19}}]
[y,edge label={node[midway]{2.32}}]
[1,edge label={node[midway,right]{\textbf{0.35}}}, edge={ultra thick}]]]
[log,edge label={node[midway,right=2pt]{-3.71}}
[sin][exp][log]]]
\end{forest}
    \caption{Starting from the root, we recursively select the node with minimum magnitude till we encounter the leaf. The path followed is root$\rightarrow$exp$\rightarrow$log$\rightarrow$1. Now, we prune the leaf and assume after fine-tuning, the loss of newer architecture is at par with the original. Thus, we accept the prune.}
    \label{fig:Eg1}
\end{figure}
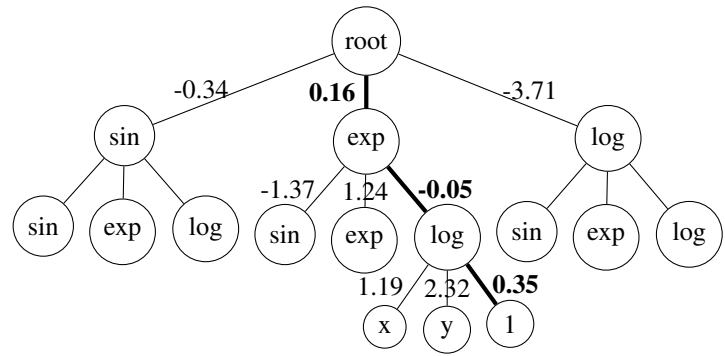

\begin{figure}[h]
    \centering
    \begin{forest}
for tree={
circle,
draw,
minimum size=7mm,
level distance=500mm,
sibling distance=500mm
}
[root
[sin,edge label={node[midway,left=2pt]{-0.37}}
[sin][exp][log]]
[exp,edge label={node[midway,left]{\textbf{0.14}}}, edge={ultra thick}
[sin,edge label={node[midway,left]{-1.29}}]
[exp,edge label={node[midway]{1.36}}]
[log,edge label={node[midway,right]{\textbf{-0.07}}}, edge={ultra thick}
[x,edge label={node[midway,left]{\textbf{0.74}}}, , edge={ultra thick}]
[y,edge label={node[midway,right]{1.86}}]
]]
[log,edge label={node[midway,right=2pt]{-3.64}}
[sin][exp][log]]]
\end{forest}
    \caption{Currently, we are at node root$\rightarrow$exp$\rightarrow$log. The node $x$ is selected and tested for pruning. Let's assume the pruning is accepted. The resulting tree is as follows:}
    \label{fig:Eg2}
\end{figure}
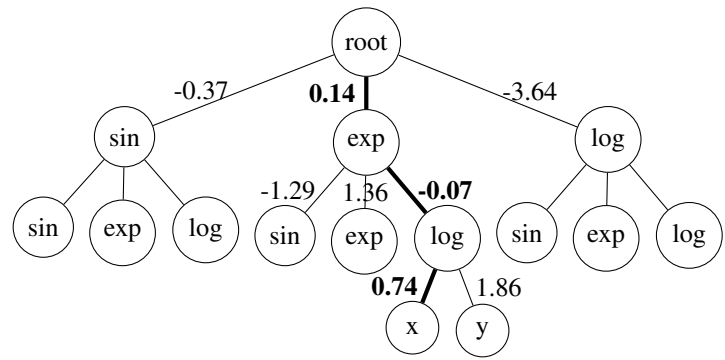

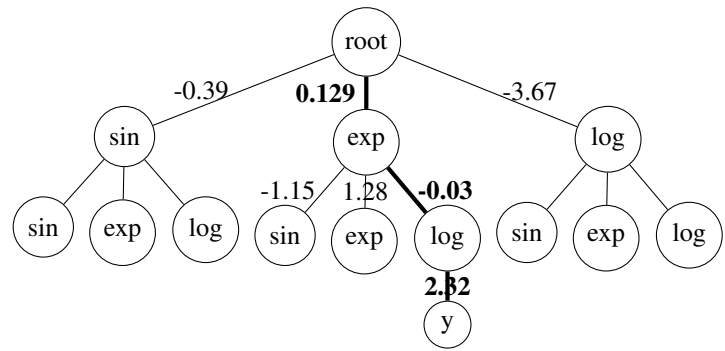
\begin{figure}[h]
    \centering
    \begin{forest}
for tree={
circle,
draw,
minimum size=2mm,
level distance=500mm,
sibling distance=500mm
}
[root
[sin,edge label={node[midway,left=2pt]{-0.39}}
[sin][exp][log]]
[exp,edge label={node[midway,left]{\textbf{0.129}}}, edge={ultra thick}
[sin,edge label={node[midway,left]{-1.15}}]
[exp,edge label={node[midway]{1.28}}]
[log,edge label={node[midway,right]{\textbf{-0.03}}}, edge={ultra thick}
[y,edge label={node[midway]{\textbf{2.32}}},, edge={ultra thick}]
]]
[log,edge label={node[midway,right=2pt]{-3.67}}
[sin][exp][log]]]
\end{forest}
    \caption{Now the node root$\rightarrow$exp$\rightarrow$log$\rightarrow$y is tested. Let's assume fine-tuned weight performance is worse, hence it's not prunable. Resulting DPA:}
    \label{fig:Eg3}
\end{figure}

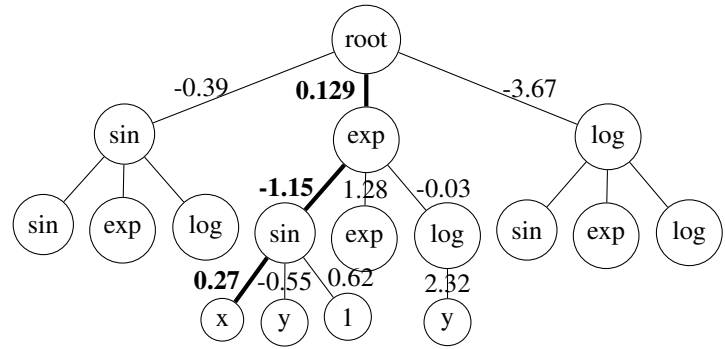
\begin{figure}[h]
    \centering
    \begin{forest}
for tree={
circle,
draw,
minimum size=2mm,
level distance=500mm,
sibling distance=500mm
}
[root
[sin,edge label={node[midway,left=2pt]{-0.39}}
[sin][exp][log]]
[exp,edge label={node[midway,left]{\textbf{0.129}}}, edge={ultra thick}
[sin,edge label={node[midway,left]{\textbf{-1.15}}}, edge={ultra thick}
[x,edge label={node[midway,left]{\textbf{0.27}}}, edge={ultra thick}]
[y,edge label={node[midway]{{-0.55}}}]
[1,edge label={node[midway,right]{{0.62}}}]]
[exp,edge label={node[midway]{1.28}}]
[log,edge label={node[midway,right]{{-0.03}}}
[y,edge label={node[midway]{{2.32}}}]
]]
[log,edge label={node[midway,right=2pt]{-3.67}}
[sin][exp][log]]]
\end{forest}
    \caption{Let's assume root$\rightarrow$exp$\rightarrow$log$\rightarrow$y is not prunable, the algorithm recurses back to root$\rightarrow$exp$\rightarrow$log. The prune is rejected again, and recurses to node root$\rightarrow$exp. The nodes selected following the algorithm are sin$\rightarrow$x, and the algorithm continues recursively, in a depth-first manner. Let's assume after all the nodes are visited, the DPA obtained is:}
    \label{fig:Eg4}
\end{figure}

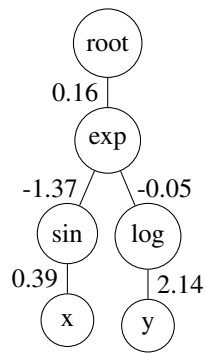
\begin{figure}[h]
    \centering
    \begin{forest}
for tree={
circle,
draw,
minimum size=7mm,
level distance=500mm,
sibling distance=500mm
}
[root
[exp,edge label={node[midway,left]{{0.16}}}
[sin,edge label={node[midway,left]{-1.37}}
[x,edge label={node[midway,left]{0.39}}]]
[log,edge label={node[midway,right]{{-0.05}}}
[y,edge label={node[midway,right]{2.14}}]
]]]
\end{forest}
    \caption{The final symbolic expression is $0.16exp(-1.37sin(0.39x)-0.05log(2.14y))$.}
    \label{fig:Eg5}
\end{figure}

\end{document}